\documentclass[runningheads]{llncs}
\usepackage[T1]{fontenc}
\usepackage{graphicx}
\usepackage{hyperref}       % hyperlinks
\usepackage{url}            % simple URL typesetting
\usepackage{booktabs}       % professional-quality tables
\usepackage{amsfonts}       % blackboard math symbols
\usepackage{nicefrac}       % compact symbols for 1/2, etc.
\usepackage{microtype}      % microtypography
\usepackage{xcolor}         % colors
\usepackage{graphicx}
\usepackage{amsmath}
\usepackage{array}
\usepackage{appendix}
\usepackage[misc]{ifsym}
\begin{document}
\title{Double Machine Learning at Scale to Predict Causal Impact of Customer Actions}
\author{Sushant More\orcidID{0000-0002-3746-2431}{\Letter} \and
Priya Kotwal\orcidID{0009-0004-6599-359X} \and
Sujith Chappidi\orcidID{0009-0009-3310-6067} \and
Dinesh Mandalapu\orcidID{0009-0007-2984-859X} \and 
Chris Khawand\orcidID{0009-0000-5283-9391}
}
\authorrunning{S. More et al.}
\institute{Amazon, Seattle WA, USA \\
\email{\{morsusha,kotwalp,jcchappi,mandalap,khawandc\}@amazon.com}
}

\maketitle              % typeset the header of the contribution
\begin{abstract}
Causal Impact (CI) of customer actions are broadly used across the industry to inform both short- and long-term investment decisions of various types.  In this paper, we apply the double machine learning (DML)
methodology to estimate the CI values across 100s of customer actions of business interest and 100s of millions of customers.    
We operationalize DML through a causal ML
library based on Spark with a flexible, JSON-driven model configuration approach
to estimate CI at scale (i.e., across hundred of actions and millions of customers). We
outline the DML methodology and implementation, and associated benefits over
the traditional potential outcomes based CI model. We show population-level as well as customer-level CI
values along with confidence intervals.  The validation metrics show a $2.2\%$ gain over the baseline methods and a $2.5X$ gain in the computational time. Our contribution is to advance the scalable
application of CI,  while also providing an interface that allows faster
experimentation,  cross-platform support,  ability to onboard new use cases,  and
improves accessibility of underlying code for partner teams. 

\end{abstract}
\section{Introduction}

Causal Impact (CI) is a measure of the incremental change in a customer's outcomes (usually spend or profit) from a customer event or action (e.g, signing up for a paid membership).  CI values are used across the industry as important signals of long-term value for multiple decisions,  such as marketing content ranking to long-term investment decisions.

Business teams are typically interested in calculating CI values for relevant actions that a customer participates in.  Some examples include customer actions such as `first purchase in category $X$',  `first $Y$ stream',  or `sign up for program $Z$'\footnote{We use placeholder $X$,$Y$,$Z$ to maintain business confidentiality}. 
The CI values are leveraged by partner teams to understand and improve the value they generate.   For many of these customer actions,  we are unable to conduct A/B experiments due to practical or legal constraints.  CI values are thus estimated off of observational data, effectively leveraging rich customer data to isolate causal relationships in the absence of a randomized experiment.

In this paper, we provide results for average treatment effects and conditional average treatment effects (i.e., customer-level CI values) estimated using a variant on the Double Machine Learning (DML) methodology \cite{Chernozukov}. The paper is arranged as follows. In Sec.~\ref{sec:ci_industry}, we give a brief overview of the use and scale of CI in the industry.  In Sec.~\ref{sec:ci_PO}, we introduce the traditional system used for calculating CI values. We also discuss the shortcomings of the traditional method and the advantages of moving to a DML-based method for calculating CI. 

Sec.~\ref{sec:dml_methodology} covers the details of our DML implementation for calculating CI values. Our contributions include improving the robustness of CI estimates through inverse propensity weighting, adding the ability to produce heterogeneous CI values, implementing customer-level confidence intervals with various assumptions,  and making available the JSON Machine Learning interface to accelerate experimentation.  We present results in Sec.~\ref{sec:results} for a few customer actions and conclude with the takeaways and ideas for future work in Sec.~\ref{sec:conclusion}.

\section{Causal impact estimation in industry}
\label{sec:ci_industry}

Causal impact estimation drives a large number of business decisions across industry.  This includes multiple organizations such as retail, search,  devices,  streaming services, and operations. To this end,  most companies have invested in developing and deploying models that vend CI values for the cutsomer actions under consideration.  In the next section, we give an overview of the traditional potential-outcome based model which is widely used in the industry for CI estimation.  This will be the baseline model.   

\subsection{CI: Potential Outcomes framework}
\label{sec:ci_PO}

CI framework applies the principles of observational causal inference. We rely on it because A/B testing is not possible to evaluate the impact of certain treatments due to practical constraints (e.g., the treatment is not effectively assignable, or would be too expensive to assign at scale). Observational causal inference methods rely on eliminating potential confounders through adjustment on observed variables. Under a "selection on observables" assumption, we believe we can estimate the causal effect correctly on average.  Applied to the customer's next 365 days of spending, for example, the CI value represents the incremental spending that a customer makes because of participating in a certain action compared to the counterfactual case where they didn’t take that particular action. The formal framework for this kind of counterfactual reasoning is the "potential outcomes" framework, sometimes known as the Neyman-Rubin causal framework \cite{Holland},  \cite{Neyman}, \cite{Rubin}. 

The potential outcome based CI model has two parts:
\begin{enumerate}
\item
Propensity binning. Group the customer based on their propensity to participate in the action. This is done based on features that relate to recency, frequency, and the monetary behavior of customers along with their other characteristics such as their tenure type.
\item
Regression adjustment. In each of the groups, we build a regression model on the control customers with customer-spend as the target. The trained model is applied on the treatment customers to predict the counterfactual spend (how much would customer have spent if they didn’t participate in the action). The difference between the predicted counterfactual and the actual spend is the CI value. We take a weighted average across different groups to get the final CI value for the customer action. 
\end{enumerate}

In addition, we require the CI model to be able to scale to the business use case.  For instance,  we may want generate CI values for hundreds of customer actions in an automated way.  In the rest of the paper, we refer to the traditional potential outcome based CI model as “CI-PO” and the DML-based model as  “CI-DML”.  

\section{CI: DML framework}

\label{sec:dml_methodology}

Note that one of the challenges in validating the causal estimates is posed by the \emph{Fundamental Problem of Causal Inference} \cite{Jasjeet}. The lack of observable ground truth makes it difficult to validate the output of a causal model. We therefore prefer models with proven desirable theoretical properties such as $\sqrt{n}$ convergence. 

The Double/Debiased Machine learning (DML) method proposed by Chernozhukov et al. \cite{Chernozukov} leverages the predictive power of modern Machine Learning (ML) methods in a principled causal estimation framework that is free of regularization bias asymptotically. 

For treatment $D$, features $X$, we express the outcome $Y$ as an additively separable function of $D$ and arbitrary function of features $X$:
\begin{equation}
Y = D \beta + g(X) + \epsilon
\label{eq:main_ci_eq}
\end{equation}
DML’s estimation strategy is motivated by writing out the residualized representation of Eq.~\eqref{eq:main_ci_eq} and its parts:
\begin{eqnarray}
\tilde{Y} &=& Y - E(Y|X) \\
\label{eq:outcome_eq}
\tilde{D}& =& D - E(D|X) \\
\label{eq:propensity_eq}
\tilde{Y} &=& \tilde{D} \beta + \tilde{\epsilon} 
\label{eq:residual_eq}
\end{eqnarray}
We use ML models to estimate $E(Y|X)$ and $E(D|X)$. The residuals from outcome equation (Eq.~\eqref{eq:outcome_eq}) are regressed on residuals from propensity equation (Eq.~\eqref{eq:propensity_eq}) to obtain the causal parameter $\beta$.  We use ML models to predict $E(Y|X)$ and $E(D|X)$.  We leverage K-fold sample splitting so that training and scoring of the ML models happens on different folds. We use a 3-fold sample split and follow the “DML2” approach \cite{Chernozukov} where we pool the residuals outcome and propensity residuals across all the folds to fit a single, final regression of the residualized outcome on the residualized treatment (Eq.~\eqref{eq:residual_eq}).

\subsection{Inverse Propensity Treatment Weighting}
\label{sec:ipw}

We use a weighted ordinary least squares to solve the residual regression equation (Eq.~\eqref{eq:residual_eq}). where the weights are determined by the Inverse Propensity Treatment Weighting (IPTW or IPW) \cite{biometric}. Our IPTW weights correspond to the Horvitz-Thompson (HT) weight \cite{Ruth},  in which the weight for each unit is the inverse of the probability of that unit being assigned to the observed group.
In Table~\ref{tab:IPW_weights} we define the weights that balance the distributions of covariates between comparison groups for two widely used estimands, the Average Treatment Effect (ATE) and Average Treatment Effect on the Treated (ATT). Weighting helps achieve additional robustness, bringing us closer to a conventional Doubly Robust estimator.  Applying these weights when conducting statistical tests or regression models helps reduce impact of confounders over and above what we get from the regression adjustment \cite{Austin}. Secondly, the weights allow us to target the estimand; the ATT has long been the preferred estimand for CI values,  since it represents the treatment effects for those customers actually treated historically.  (We refer to the customer-level counterparts of ATE, ATT estimands as HTE and HTT respectively.)
%\begin{center}
\begin{table}
\caption{IPW weights for different estimands.  $D$ is the treatment assignment and $\hat{e}(X)$ is the treatment propensity,  $E(D|X)$.}
\label{tab:IPW_weights}
\begin{tabular}{|m{6cm}|m{4cm}|}
\hline
Estimand & IPW weight \\
\hline
\hline
Average treatment effect (ATE) & $\dfrac{D}{\hat{e}(X)} + \dfrac{1-D}{1 - \hat{e}(X)}$ \\
\hline
Average treatment effect on the treated (ATT) & $D +(1-D) \dfrac{\hat{e}(X)}{1 - \hat{e}(X)}$ \\
\hline
\end{tabular}
\end{table}
%\end{center}
%

\subsection{Common support and propensity score trimming}
\label{subsec:common_support}

\begin{figure}[h]
  \centering
  \includegraphics[width=0.65\linewidth]{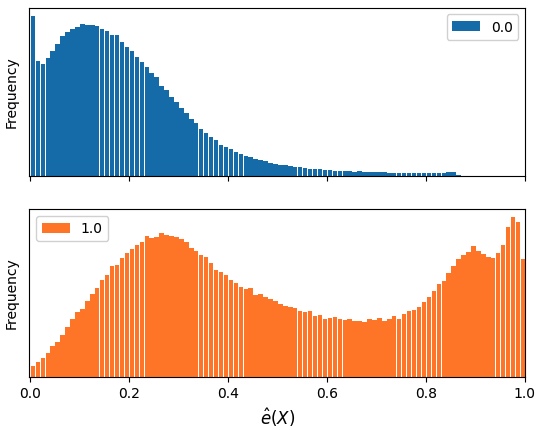}
  \caption{Representiative propensity scores distribution for control (top panel) and treatment (bottom panel) groups.}
  \label{fig:propensity_scores_dist}
\end{figure}

For many treatments, propensity distribution has significant mass near 1 for the treated group and near 0 for the control group (see an example histogram in Fig.~\ref{fig:propensity_scores_dist}).  Scores near the boundary can create instability in weighting methods.  In addition, these scores often represent units for whom we cannot make an adequate treatment-control comparison. We limit analysis to the common support region, where propensity score distributions overlap between treated and untreated samples. 

We also use trimming to exclude customers whose estimated propensity is outside of the range $[\alpha, 1-\alpha]$.  We experimented with different thresholds on various customer actions and observed that $\alpha=0.001$ with rescaled propensity scores works the best.

\subsubsection{Normalizing and rescaling weights}

\begin{figure}
  \centering
  \includegraphics[width=0.75\linewidth]{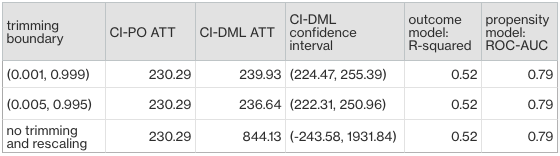}
  \caption{Effect of propensity scores trimming and rescaling on estimated CI for a certain customer action.}
  \label{fig:propensity_trimmed}
\end{figure}

When using the IPW, we normalize the weights by rescaling the propensity scores for each customer $i$ as in Eq.~\eqref{eq:scaled_propensity}.
\begin{equation}
\hat{e}(X_i)_{{\rm scaled}} = \left(\dfrac{\overline{D}}{\overline{\hat{e}(X)}}\right) \ast \hat{e}(X_i)
\label{eq:scaled_propensity}
\end{equation}
$\overline{D}$ and $\overline{\hat{e}(X)}$ in Eq.~\eqref{eq:scaled_propensity} are the averages of treatment assignment and propensity score respectively taken over both the treatment and control population combined. 
Propensity trimming and rescaling reduces variance, leads to more stable estimates, and tighter confidence intervals as seen in Fig.~\ref{fig:propensity_trimmed}. 

\subsection{Heterogeneity in DML}
CI-DML implements a version of the heterogenous effects modeling proposed in \cite{Cherno_goldman_semenova_taddy}, by leveraging the treatment-feature interactions in the final stage of DML to identify heterogenous (customer-level) responses. 
The general form of Eq.~\eqref{eq:residual_eq} can be written as 
\begin{equation}
\tilde{Y} = h(X, \tilde{D}) + \tilde{\epsilon}\;.
\label{eq:res_het}
\end{equation} 
In fact, Eq.~\eqref{eq:residual_eq} is a special case of Eq.~\eqref{eq:res_het} with $h(X, \tilde{D}) = \tilde{D} \beta$.  We interact treatment with the features and define $h(X, \tilde{D}) \equiv \psi(X) * \tilde{D} \, \beta$,  where `$\ast$' represents element-wise multiplication. Thus, the heterogeneous residual regression becomes:
\begin{equation}
\tilde{Y} = \psi(X) * \tilde{D} \, \beta + \tilde{\epsilon}
\label{eq:het_residual_eq}
\end{equation}
We want $\psi(X)$ to be low-dimensional so that we are able to extract the coefficient $\beta$ in Eq.~\eqref{eq:het_residual_eq} reliably. 

\begin{figure}[h!]
  \centering
  \includegraphics[width=0.6\linewidth]{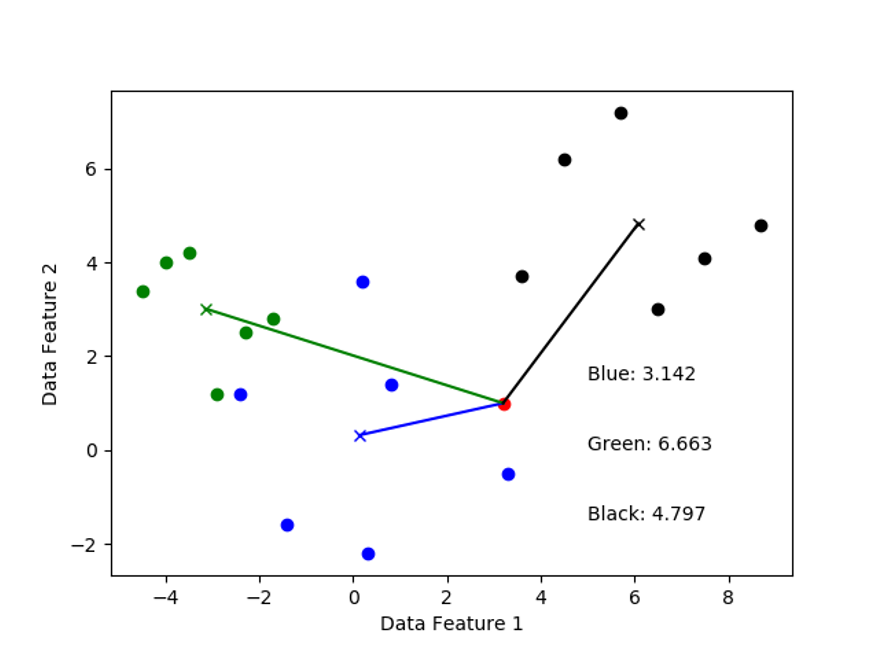}
  \caption{Schematic for calculation of distance from cluster centroids. The red dot is represented by three features which is the distance from centroids from blue, green, and black clusters. }
  \label{fig:cluster_distance}
\end{figure}

Let $N$ and $M$ be the number of customers and features respectively. If the dimension of $\psi(X)$ is $N\times K$, we want $K \ll M$. In our use case,  $M$ is typically ~2000 and $K$ is typically around 20. To get the low-dimensional representation, $\psi(X)$ we proceed as follows: 
\begin{enumerate}
\item
We project the original features onto an orthogonal space through Principal Component Analysis (PCA). 
\item
We run a K-means clustering algorithm on the highest-signal Principal Components. Dimension reduction from PCA helps to reduce dimensionality-related problems when computing Euclidean distance for K-means clustering.
\item
We calculate the $K$ cluster scores for each customer, as $\psi_{i,c} = \dfrac{1/d_{i,c}}{\sum_{k=1}^K 1/d_{i,k}}$ where $d_{i,c}$ is the distance of customer $i$'s value from centroid of cluster $c$ (see schematic in Fig.~\ref{fig:cluster_distance}).
\end{enumerate}

Once we have calculated the distance features $\psi(X)$ for each customer, we interact them with the propensity residuals and fit a linear regression model using IPW (refer sec.~\ref{sec:ipw}) to extract the coefficients $\beta$ in Eq.~\eqref{eq:het_residual_eq}. The heterogenous estimates are given by 
\begin{equation}
h = \psi(X) \beta \;.
\label{eq:hetero_ci}
\end{equation}
E.g., for $K=3$,  $h = \psi_1 \beta_1 +\psi_2 \beta_2 + \psi_3 \beta_3$. 

\begin{figure}[h!]
  \centering
  \includegraphics[width=\linewidth]{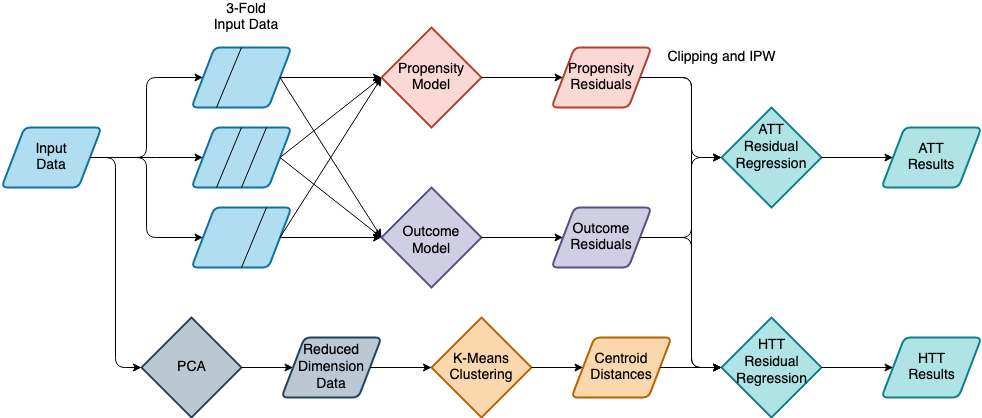}
  \caption{Schematic of the CI-DML modeling framework.}
  \label{fig:ci_DML}
\end{figure}

A schematic of CI-DML workflow is shown in Fig.~\ref{fig:ci_DML}.

\subsection{Confidence intervals in DML}

One of the disadvantages of CI-PO is that generating confidence intervals requires bootstrapping around the multi-step process and is computationally expensive.  Obtaining confidence intervals in CI-DML is straightforward. For a single ATT parameter estimate, we obtain the confidence interval simply by calculating the variance of the estimate of $\beta$ in Eq.~\eqref{eq:residual_eq}. We also estimate Huber-White heteroscedasticity consistent standard errors \cite{Huber}, \cite{White}. For the ATT case, the steps for calculating variance of the coefficient $\hat{\beta}$ are as follows:
\begin{equation}
Var(\beta) = H \hat{\Sigma} H' .
\label{eq:Var}
\end{equation}
For a customer, `$p$':
\begin{equation}
\hat{\sigma}_p^2 = \hat{U}_p^2 = (\tilde{Y}_p - \tilde{D}_p \hat{\beta})' (\tilde{Y}_p - \tilde{D}_p \hat{\beta})\;,
\label{eq:sigma_p_att}
\end{equation}
where $\hat{\beta}$ is the value of coefficient from solving Eq.~\eqref{eq:residual_eq}.  Note that $\tilde{Y}_p$ and $\tilde{D}_p$ are scalars. 
$\Sigma$ in Eq.~\eqref{eq:Var} is a diagonal matrix with the squared prediction error $\hat{\sigma}_p^2$  for each customer on its diagonal and $H$ in Eq.~\eqref{eq:Var} is defined as 
\begin{equation}
H = (\tilde{D}' \ast W \tilde{D})^{-1} \tilde{D}' \ast W
\label{eq:H_def_att}
\end{equation}
where $W$ are the IPW weights as defined in Sec.~\ref{sec:ipw}. 

We compute the confidence intervals on the causal estimate $\beta$ using $Var(\beta)$.

\subsubsection{Customer-level confidence intervals}

CI-DML also provides the ability to obtain customer-level confidence intervals. From Eq.~\eqref{eq:hetero_ci},  we can write
\begin{equation}
Var(h) = Var(\psi \beta) =  \sum_k \psi_k^2 Var(\beta_k) + \sum_{k\neq l} \psi_k \psi_l Cov(\beta_k, \beta_l).
\label{eq:htt_variance}
\end{equation}

We calculate variance of the heterogeneous coefficients following similar approach as in Eqs.~\ref{eq:Var}, \ref{eq:sigma_p_att}, and \ref{eq:H_def_att}. The only difference is we replace $\tilde{D}_p \rightarrow \psi(X_p) * \tilde{D}_p$ in Eq.~\eqref{eq:sigma_p_att} and $\tilde{D} \rightarrow \psi(X)*\tilde{D}$ in Eq.~\eqref{eq:H_def_att}. 

For the ATT case,  $Var(\beta)$ is a scalar whereas for the HTT case, $Var(\beta)$ is a $K\times K$ matrix. The first and second terms in the summation in Eq.~\eqref{eq:htt_variance} are the diagonal and the off-diagonal terms of the $Var(\beta)$ matrix respectively. 

\subsection{DML implementation}

\begin{figure}[h!]
  \centering
  \includegraphics[width=0.75\linewidth]{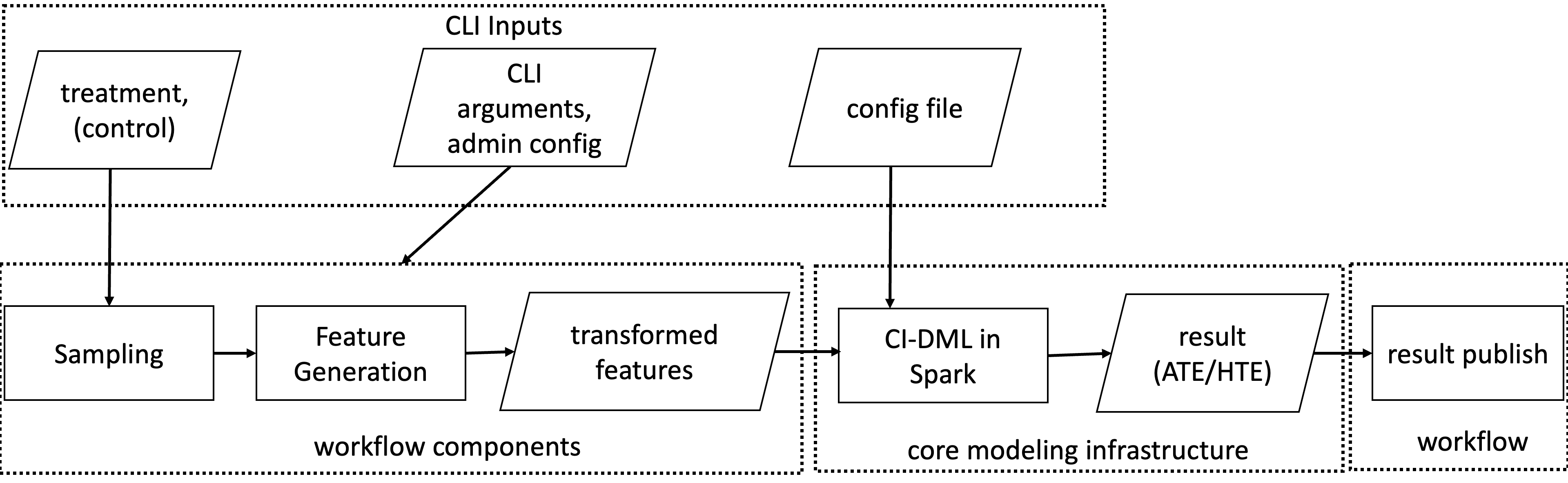}
  \caption{JSON-Machine Learning Stage Interpreter modeling stages}
   \label{fig:jmi}
\end{figure}

We developed a causal ML library with JSON driven modeling configuration (see Fig.~\ref{fig:jmi}).  JSON ML Interpreter (JMI) translates JSON configuration to executable Python ML application. 

The main advantages of JMI approach are:\\
\textbf{Flexibility}: Business questions from various domains cannot always be addressed through a single unified configuration of a causal model. We address this in CI-DML where users can invoke different causal analysis frameworks (DML,  Causal Forests) and prediction algorithm type (regression, classification, clustering). \\
\textbf{Scalability}: CI-DML utilizes distributed implementation of algorithms and file system via Apache Spark which helps causal modeling at  the big-data scale (100 millions customers,  multiple targets, and time horizons) \\
\textbf{Persistence}: CI-DML inherits SparkML serialization and deserialization methods to persist and instantiate fitted models. \\
\textbf{Compatibility}: In addition to Spark, interfaces to adapt ML libraries from scikit-learn, tensorflow, MXNet and other communities can be onboarded using the configurable abstraction support by JMI.

In the CI-DML system, we dockerize the JMI Causal ML library which is platform agnostic which has the flexibility to extend and utilize different compute engines like AWS EMR, Sagemaker or AWS Batch based on the use case and will abstract the computation information from the user.  Dockerization also helps version control and the build environment via standard software development tooling. 

\section{Results}
\label{sec:results}

Next we present the results for CI-DML. The target variables we focus on is customer spending, but our framework on can be leveraged to obtain the causal impact on any other target variable of interest (e.g., net profit, units bought etc).  For every CI run, we produce both the population-level ATT values (Eq.~\eqref{eq:residual_eq}) and the customer-level HTT (Eq.~\eqref{eq:hetero_ci}) values.

\begin{figure}[h!]
  \centering
  \includegraphics[width=\linewidth]{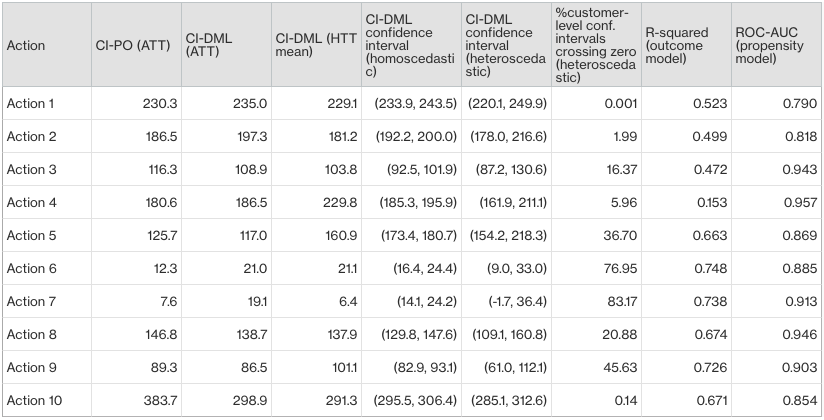}
  \caption{CI values and confidence intervals for selected customer actions.}
    \label{tab:ci_big_table}
\end{figure}

We compared the CI-PO and CI-DML results for 100+ customer actions.  As noted earlier, two major advantages of CI-DML are the availability of customer grain results (aka.  HTT) and confidence intervals.   
In Fig.~\ref{tab:ci_big_table}, we present population-level (ATT) and customer-level values for selected representative customer actions \footnote{We anonymize actions to preserve business confidentiality}.  
The reported confidence intervals are for both homoscedastic and heteroscedastic error variances.  To get a sense of the level of variance in customer-grain results, we report the percentage of customers where the  customer-level confidence interval crosses zero. We also report the out-of-sample fit metrics for outcome and propensity model in DML. 

Takeaways from analysis of 100+ actions are:
\begin{enumerate}
\item
Population-level CI-PO and CI-DML values are aligned for 86\% of actions. 
\item
When the customer-level CI values are aggregated up, they are generally aligned with the population-level CI-DML values. 
\item
The difference between the CI-PO and CI-DML values are larger either when the data is noisy and/or we have a small sample size.  For such cases, we also see large confidence intervals and the mean of HTT values is farther away from the CI-DML ATT values.
\item
The homoscedastic confidence intervals are tighter than the heteroscedastic confidence interval as expected.  However, the homoscedastic confidence intervals likely under-predict the variance. We recommend business stakeholders to use the heteroscedastic results while making decisions. 
\item
The customer-level confidence intervals are economically reasonable.  The percentage of customer-level confidence interval crossing zero increases for data with lower participation and is small for customer actions with a long history. 
\item
The ML model metrics shown in Fig.~\ref{tab:ci_big_table} are using ridge regression for the outcome model and logistic regression for the propensity model. We noticed that the model metrics as well as the CI values are relatively insensitive to the choice of ML model at the outcome/propensity stage.  Accordingly, we leverage ridge and logistic models due to their favorable compute time.
\end{enumerate}

\subsection{Hyperparameter tuning}

The hyperparameters (e.g., regularization strength) in the outcome and propensity model are chosen based on the out-of-sample performance.  
For the HTT estimates, the two main hyperparameters are the number of principal components and the number of clusters.  

\begin{figure}[h!]
  \centering
  \includegraphics[width=0.6\linewidth]{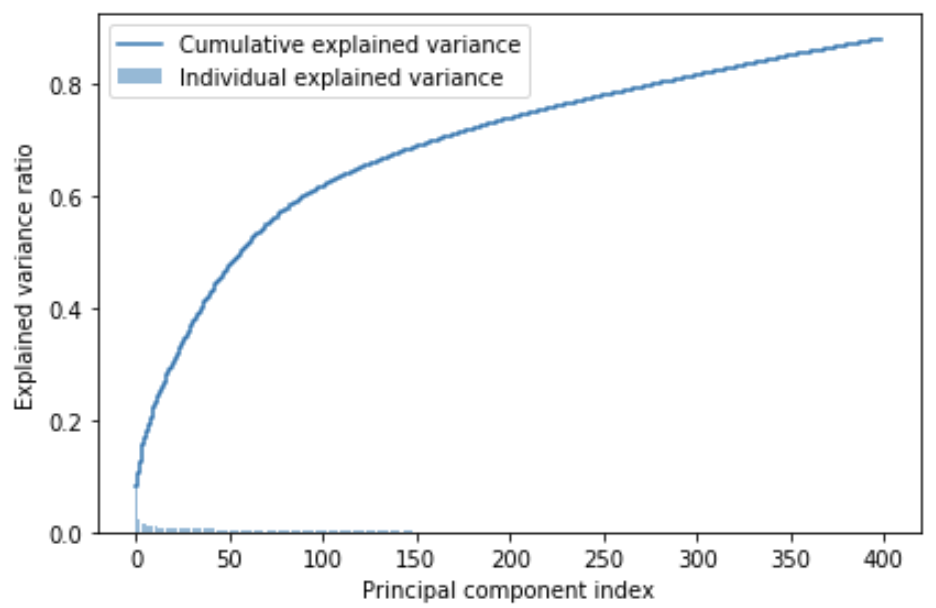}
  \caption{Amount of variance explained as a function of principal components. }
  \label{fig:pc_variance_explained}
\end{figure}

We choose the number of principal components (PC) based on the percentage of variance explained.  We find that around 300 PC, about $80\%$ of the variance is explained (Fig.~\ref{fig:pc_variance_explained}). The amount of variance explained grows much slowly as we add more number of PC. To avoid sparsity issues in the downstream K-means calculation, we choose the number of PC components to be 300. 

Choosing the number of clusters is less straightforward. Standard tools such as elbow method and Silhouette score do not yield a clear answer for the optimal cluster choice.  In the current work, we choose 20 clusters as we think its best suited for our data. One of our ongoing work, is to make this choice in a more data-driven way.  We also find that the mean of HTT values is robust with respect to the choice for number of clusters. 

\subsection{Spread of customer-level CI values}

\begin{figure}[h!]
  \centering
  \includegraphics[width=0.6\linewidth]{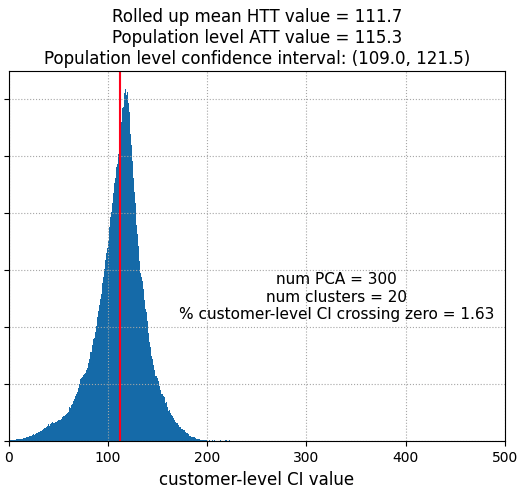}
  \caption{Spread of customer-level CI values.  The red line is the mean of customer-level CI values. }
   \label{fig:customer_level_ci_hgram}
\end{figure}

So far, we have only looked at the mean of customer-level values in Fig~\ref{tab:ci_big_table}.  In Fig.~\ref{fig:customer_level_ci_hgram}, we look at the customer-level CI scores for a representative customer action.  We see that most of the customers have CI value close to the average HTT value. We see that there are few customers with a low CI value which shifts the mean to the left.  One of the industry application of CI is marketing optimization. Having access to distribution of customer-level CI scores as in Fig.~\ref{fig:customer_level_ci_hgram} can help personalization and aid finer decision making. 

\section{Validation}

\subsection{Placebo tests}
\begin{figure}[h!]
  \centering
  \includegraphics[width=0.6\linewidth]{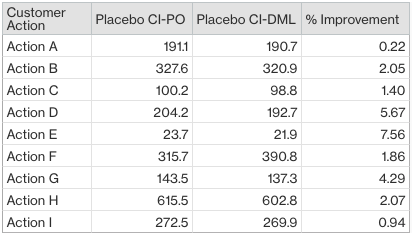}
  \caption{Placebo results for selected customer actions for CI-PO and CI-DML model. }
   \label{fig:placebo_results}
\end{figure}

Placebo tests help us understand the relative ability of competing causal estimates to account for selection bias. Selection bias occurs
when customers who take an action (e.g., stream video) have unobserved characteristics not included in the model that makes them
systematically more or less likely to take the action (e.g., high income, low age etc.). In placebo tests, we take the treatment group
customers and simulate as though they took the action a year before the actual date. This is achieved by shifting the event date by one year and recalculating the features based on the shifted event date. The CI estimated in this set up is the
“placebo error”.  Since, this is a fake event, a model with a lower placebo error than another suggests that it has a lower selection bias.  Running placebo tests on all events is computationally expensive,  so we selected a few events for placebo analysis.  The results are shown in Fig.~\ref{fig:placebo_results}. 

The key takeaways from Fig.~\ref{fig:placebo_results} are:
\begin{itemize}
\item
Selection bias is inherently event-dependent. When averaged across the selected customer actions,  we see a $2.24\%$ improvement in placebo estimates when going from CI-PO to CI-DML. 
\item
Selection bias is primarily impacted by the modeling features.  As CI-PO and CI-DML use the same features, we did not expect
big improvements in placebo tests. The consistent improvement across the events shows that double machine learning
methodology is better able to adjust for observables even when the same features are used. 
\end{itemize} 

\subsection{Confidence interval comparison}

\begin{figure}[h!]
  \centering
  \includegraphics[width=0.6\linewidth]{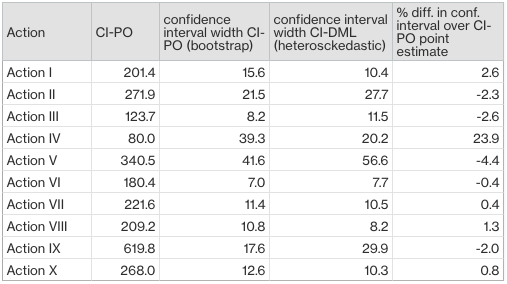}
  \caption{CI-PO bootstrap vs. CI-DML confidence interval width comparison}
   \label{fig:bootstrap_comparison}
\end{figure}

One of the major wins in CI-DML is that we provide heteroskedasticity-consistent confidence intervals at both a customer and
aggregate level for every CI analysis in a scalable and lower-cost fashion. We compare the uncertainty estimates (specifically the
width of confidence intervals) from CI-DML with the bootstrap results in CI-PO for a few events in Fig.~\ref{fig:bootstrap_comparison}.

We find confidence interval width to be comparable among the two approaches. On average, the CI-DML width (scaled with CI-PO point
estimate) is 1.5\% smaller when compared to bootstrap-based confidence interval.  A bootstrap-enabled CI-PO run takes about 2.5X more time than a CI-DML run.  Bootstrap also does not scale for events with large number of customers.  As CI-DML approach for confidence intervals is based on a closed form implementation,  we do not have any scalability issues. In addition, note that bootstrapping has theoretical limitations when used for matching estimators \cite{Abadie_Imbens}.

\section{Conclusion and Future work}
\label{sec:conclusion}
In this work, we introduced a state-of-the-art methodology used for calculating CI values.  We noted that a DML based framework eliminates bias, allows us to extract heterogeneity in CI values, and provides a scalable way to construct heteroscedastic confidence intervals. We also made a case for using IPW and common support to refine the CI estimates. We demonstrated how leveraging PCA followed by K-means clustering allowed us to introduce customer-level heterogeneity. Using JSON based config allows flexibility to experiment with a wide variety algorithms and can take us from experimentation to production in minimal steps. 

We presented results for few anonymized customer actions across different domains. Both the population-level and customer-level results for the customer actions we have looked at so far are aligned with the CI-PO results and our expectations. 

Note that estimation of heterogeneous or context-aware treatment effects is an active area of research with wide applications ranging from marketing to health care.  Distribution of treatment effects across different subgroups, or as a function of specific individual-level characteristics provides researchers with additional insights about the treatment/ intervention analyzed.  Our work showcases, a scalable real-world application for extracting average as well heterogeneous causal estimates which we believe will be of interest to the broader scientific community. 

\subsection{Future work}

Validating the causal estimates is challenging due to lack of ground truth. In the current work, we relied on the model fit metrics in the DML steps,  placebo tests,  and on bridging the CI-DML and CI-PO outputs.  
In the future, we plan to include metrics which focus on the validation of heterogeneous treatment effects.  Examples of these include metrics based on Generic Machine Learning \cite{GLM} and empirically calibrated Monte Carlo resampling techniques \cite{MC}.
We also plan to experiment with doubly robust estimator in the DML stage. 

%%%%%
%%%%%%


\newpage

\begin{thebibliography}{999}

\bibliographystyle{plain}

\bibitem{Chernozukov}
Victor Chernozhukov, Denis Chetverikov, Mert Demirer, Esther Duflo, Christian Hansen, Whitney Newey, James Robins.  Double/debiased machine learning for treatment and structural parameters. \emph{The Econometrics Journal},  Volume 21, Issue 1, 1 February 2018,  Pages C1–C68,  \href{https://doi.org/10.1111/ectj.12097}{doi.org/10.1111/ectj.12097}

\bibitem{Jasjeet}
Sekhon, Jasjeet.  The Neyman–Rubin Model of Causal Inference and Estimation via Matching Methods.  2007. \emph{The Oxford Handbook of Political Methodology.}

\bibitem{Holland}
Holland, Paul W.  Statistics and Causal Inference.  1986.  \emph{J. Amer. Statist. Assoc. } 81 (396): 945–960.  \href{https://doi.org/10.1080/01621459.1986.10478354}{doi:10.1080/01621459.1986.10478354}

\bibitem{Neyman}
Neyman, Jerzy.  Sur les applications de la theorie des probabilites aux experiences agricoles: Essai des principes. Master's Thesis (1923).  Excerpts reprinted in English,  \emph{Statistical Science}, Vol. 5, pp. 463–472.

\bibitem{Rubin}
Rubin, Donald.  Causal Inference Using Potential Outcomes. 2005.  \emph{J. Amer. Statist. Assoc. } 81 (396): 945–960.  \href{https://doi.org/10.1080/01621459.1986.10478354}{doi:10.1080/01621459.1986.10478354}

\bibitem{Cherno_goldman_semenova_taddy}
Chernozhukov, V., Goldman, M., Semenova, V.,  and Taddy, M. (2017). Orthogonal Machine Learning for Demand
Estimation: High Dimensional Causal Inference in Dynamic Panels. ArXiv:1712.09988 [Stat]. 

\bibitem{Huber}
Huber, Peter J.   The behavior of maximum likelihood estimates under nonstandard conditions. (1967) \emph{Proceedings of the Fifth Berkeley Symposium on Mathematical Statistics and Probability.} 
Vol. 5. pp. 221–233

\bibitem{White}
White,  Halbert.  A Heteroskedasticity-Consistent Covariance Matrix Estimator and a Direct Test for Heteroskedasticity.  \emph{ Econometrica} 48 (4): 817–838

\bibitem{HTR1}
Horvitz, D. G.,  and Thompson, D. J.  . A Generalization of Sampling Without Re-placement From a Finite Universe (1952).  \emph{Journal of the American Statistical Association},  47(260), 663-685

\bibitem{HTR2}
Nie, X., and Wager, S.  Quasi-Oracle Estimation of Heterogeneous Treatment Effects (2017).  ArXiv:1712.04912 [Econ, Math, Stat]

\bibitem{DR}
Edward H. Kennedy,  Optimal doubly robust estimation of heterogeneous causal effects (2020).  ArXiv:2004.14497 [math.ST]

\bibitem{biometric}
Estimating exposure effects by modelling the expectation of exposure conditional on confounders. Biometrics 48 479–495. MR1173493

\bibitem{Ruth}
Ruth C, Brownell M, Isbister J, MacWilliam L, Gammon H, Singal D, Soodeen R, McGowan K, Kulbaba C, Boriskewich E. Long-Term Outcomes Of Manitoba's Insight Mentoring Program: A Comparative Statistical Analysis . Winnipeg, MB: Manitoba Centre for Health Policy, 2015

\bibitem{Austin}
P. C. Austin and E. A. Stuart, Moving towards Best Practice When using Inverse Probability of Treatment Weighting (IPTW) Using the Propensity Score to Estimate Causal Treatment Effects in Observational Studies, vol. 34, pp. 3661-3679, 2015.

\bibitem{Hirano}
Hirano, K., Imbens, G.W. Estimation of Causal Effects using Propensity Score Weighting: An Application to Data on Right Heart Catheterization. Health Services and Outcomes Research Methodology 2, 259–278 (2001). \href{https://doi.org/10.1023/A:1020371312283}{doi.org/10.1023/A:1020371312283}

\bibitem{Abadie_Imbens}
A. Abadie and G. Imbens, On the failure of bootstrap for matching estimators, Econometrica, Vol. 76, No. 6 (2008), 1537-1157

\bibitem{GLM}
Chernozhukov, Victor, Mert Demirer, Esther Duflo, and Ivan Fernandez-Val (2022).
Generic machine learning inference on heterogenous treatment effects in randomized experiments.
\href{https://arxiv.org/abs/1712.04802}{aXiv:1712.04802v6 [stat.ML]}

\bibitem{MC}
Knaus, M. C., Lechner, M., \& Strittmatter, A. (2021). Machine learning estimation of heterogeneous causal effects: Empirical monte carlo evidence. The Econometrics Journal, 24(1), 134-161

\end{thebibliography}
%

%\begin{appendices}

	\appendix
\section*{Appendix}

\section{Sample JSON config} \label{sec:model_metrics}

\begin{figure}
  \centering
  \includegraphics[width=0.8\linewidth]{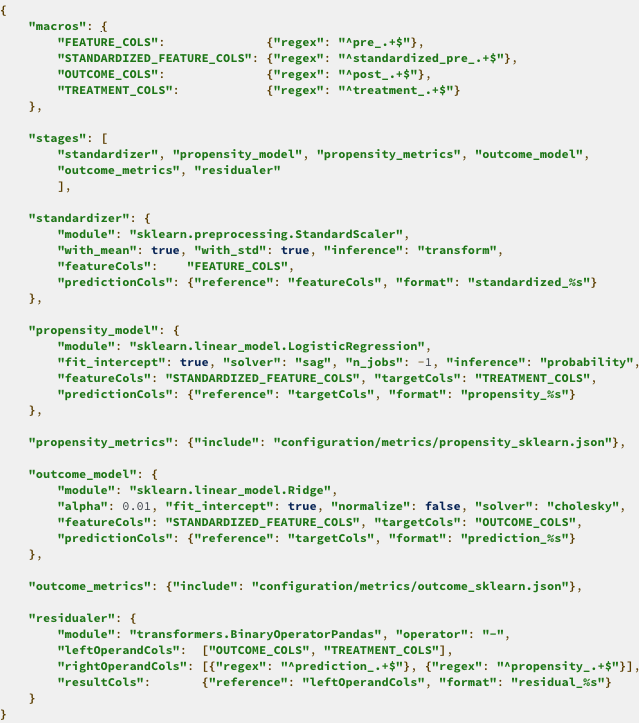}
  \caption{A sample JSON config where we are using Ridge regression for the outcome model and the logistic regression for the propensity model.}
  \label{fig:sample_json_config}
\end{figure}

We show a snippet of JSON config in Fig.~\ref{fig:sample_json_config}.   
We can swap the specified models in the outcome and propensity step with any ML model of our choice.  
Likewise we can easily configure pre/post-processing steps and hyperparameters through JSON files. 

%==================================================================

%\newpage
%==================================================================

%\end{appendices}

\end{document}